%% file: acl_latex.tex
\documentclass[11pt]{article}

\usepackage[preprint]{acl}
\usepackage{multirow}
\usepackage{times}
\usepackage{latexsym}

\usepackage[T1]{fontenc}

\usepackage[utf8]{inputenc}

\usepackage{microtype}

\usepackage{inconsolata}

\usepackage{graphicx}

\usepackage{comment}
\usepackage{amsmath}
\usepackage{amssymb}
\usepackage{bbold}
\usepackage{booktabs}
\usepackage{xcolor}

\usepackage{array}
\newcolumntype{L}[1]{>{\raggedright\arraybackslash}p{#1}}

\title{What Do LLMs Know About Alzheimer's Disease?  \\ Multi-loss Fine-Tuning and Probing for AD Detection}

\author{    
    Lei Jiang  \\
  University of Illinois Chicago  \\
  \texttt{ljian43@uic.edu} \\\And
  Yue Zhou  \\
  University of Illinois Chicago \\
  \texttt{yzhou232@uic.edu} \\\And
  Natalie Parde \\
 University of Illinois Chicago  \\
  \texttt{parde@uic.edu} \\
  }

\begin{document}
\maketitle
\begin{abstract}

Reliable early detection of Alzheimer's disease (AD) is challenging, particularly due to the limited availability of labeled data. While large language models (LLMs) have shown strong transfer capabilities across do  mains, adapting them to the AD domain through supervised fine-tuning remains largely unexplored. In this work, we empirically evaluate various model architectures across three heterogeneous transcript corpora (Pitt, CCC, ADRC) to investigate their effectiveness for text-based AD detection and analyze how task-relevant information is encoded within their internal representations. To the best of our knowledge, our fine-tuned BERT and T5 models establish a new state-of-the-art on the Pitt and CCC datasets, while achieving strong performance on ADRC. In parallel, the decoder-only Llama-1B achieves highly competitive results comparable to BERT and T5 across all three corpora, highlighting its effectiveness for AD detection. We further conduct a comprehensive evaluation of the Llama-1B backbone, analyzing cross-corpus transferability, optimal input chunk-size granularity, and the impact of clinical transcript markers. Also, we use linear probing to empirically show that fine-tuning shifts the representations of individual tokens, both linguistic markers and content words, in ways that reflect AD-related signal.

\end{abstract}

\input{1.intro}

\input{2.related}

\input{3.methodology}
\input{4.main_experiments}
\input{5.additional}
\input{6.conclusion}
\input{7.limitation}


\bibliography{custom}

\appendix

\input{8.appendix}

\end{document}

%% file: 1.intro.tex
\section{Introduction}

Alzheimer's disease (AD) leads to progressive cognitive decline and poses a major burden on patients and caregivers \cite{skaria2022economic}, and early detection of AD is increasingly crucial to enable timely intervention and improve patient outcomes as populations age globally \cite{doi:10.3233/JAD-200888}.
Recently, large language models (LLMs) have demonstrated success at a myriad of downstream tasks through fine-tuning on task-relevant datasets, allowing them to combine the wealth of knowledge learned during pre-training with more task-specific details \cite{hu2022lora,NEURIPS2023_1feb8787}. This creates promise for advanced, AI-enabled solutions to challenging problem domains, including those related to healthcare problems growing in prevalence, such as AD \cite{farzana-parde-2024-domain,han-etal-2025-llm-based,li-etal-2025-large-foundation}.

Current methods for AI-enabled AD detection rely on standardized linguistic and cognitive assessments and face challenges in accuracy and scalability \cite{Martinc2020TacklingTA,balagopalan20_interspeech,yuan20_interspeech,9414147,rohanian21_interspeech,farzana-parde-2022-interaction,farzana-parde-2023-towards}. The rapid advancement of LLMs presents opportunity to leverage their strong linguistic understanding for this clinical application, but fine-tuning language models within the AD domain is underexplored \cite{zhang-2025-llm-ad, hou-2025-llm-agent,li-2025-multi-agent-llm, Dhinagar-2025-moe}, in large part due to the limited availability of labeled clinical data \cite{duan-etal-2023-cda}. Addressing these challenges can not only enhance diagnostic tools for AD but can also contribute to broader research on how LLMs can be effectively adapted to specialized, low-resource clinical tasks without compromising their general language understanding.


Research on other language tasks has demonstrated that supervised fine-tuning (SFT) allows models to learn directly from curated examples that reflect the desired outputs, improving reliability, consistency, and alignment with domain expectations or usage goals \cite{DBLP:conf/iclr/WeiBZGYLDDL22,harada-etal-2025-massive}. This process not only enhances task performance and reduces unpredictable behavior, but also enables efficient domain adaptation, safety and policy alignment, and personalization without requiring that the model is retrained from scratch \cite{JMLR:v25:23-0870, openai2024gpt4technicalreport, ouyang2022training,peng2023instructiontuninggpt4}.

In this work, we comprehensively evaluate three language model architectures for AD detection, assessing their capacity to encode clinically meaningful linguistic markers. Within our multi-loss framework, encoder and encoder-decoder architectures such as BERT\cite{devlin2019bert} and T5\cite{raffel2023exploringlimitstransferlearning} yield superior classification performance—achieving, to the best of our knowledge, state-of-the-art results on the Pitt \cite{pitt_corpus} and CCC \cite{ccc_corpus} corpora, alongside strong performance on the SLaCAD \cite{farzana-etal-2024-slacad} dataset. Importantly, the decoder-only Llama3.2-1B-Instruct\cite{grattafiori2024llama3herdmodels}(refer to it as Llama-1B throughout the paper) achieves highly competitive results comparable to BERT and T5 across all three heterogeneous corpora, motivating further investigation into its broader diagnostic potential. While BERT and T5 excel at raw classification performance, larger decoder-only architectures offer strong contextual modeling capabilities that may benefit future clinical applications. Although exploring these capabilities falls outside the scope of this study, our extensive evaluation establishes an important baseline for future work on scalable, multi-corpus clinical AD detection workflows.

Our contributions are fourfold.
\textbf{(1)} We propose a multi-loss SFT recipe
for a Llama-1B backbone that achieves strong AD-detection performance
across three heterogeneous transcript corpora (Pitt, CCC, SLaCAD),
demonstrating that a small open-weight LLM can match or exceed prior
specialized models on this task.
\textbf{(2)} Through a per-corpus loss-term ablation, we show that all three corpora contribute complementary signal: the full joint objective
is best on every per-corpus metric, and removing any single
$\mathcal{L}_d$
degrades performance on the remaining corpora as well as on the dropped
one.
\textbf{(3)} We identify chunk-size selection as a critical
preprocessing choice: aligning the chunk size to the length scale of the shortest representative segment ($512$ tokens in our setting) gives all training examples a comparable input granularity and improves $F_1$
uniformly across corpora.
\textbf{(4)} We show that the CHAT/CLAN markers carried by the Pitt prompt are a transferable AD-domain feature.  The markers explicitly tag the
disfluencies, retraces, and word-finding pauses that other corpora leave unannotated, and retaining them improves generalization on
\emph{both} Pitt and unmarked CCC, indicating that a small but genuine AD-domain signal is extracted from Pitt and reused on inputs that lack
the markers.
\textbf{{5}} We use a layer-wise linear probing analysis on LLM representations and empirically observe that, after fine-tuning, the representations of individual tokens, including both linguistic markers and content words, shift in ways that may reflect AD-related signal.

%% file: 2.related.tex
\section{Related Work} 
\begin{table*}[t]
\centering
\small
\begin{tabular}{@{}lp{3.0cm}p{3.2cm}p{1.6cm}p{1.8cm}@{}}
\toprule
\textbf{Corpus} & \textbf{Elicitation} & \textbf{Transcription} & \textbf{Total} & \textbf{Ctrl : AD} \\
\midrule
Pitt   & Cookie Theft picture description & CHAT/CLAN with disfluency markers & 1{,}289 & 19\% : 81\% \\
CCC    & Open-domain dyadic interview     & Orthographic, no markup           &   105   & 46\% : 54\% \\
SLaCAD & Autobiographical narrative       & Orthographic, no markup           &    91   & 90\% : 10\% \\
\bottomrule
\end{tabular}
\caption{Corpora differ in genre, annotation richness, and direction of class imbalance. }
\label{tab:corpora-description}
\end{table*}
\subsection{Alzheimer's Disease Detection}

A relatively large body of work within the NLP community has studied AD detection from different angles~\cite{farzana-parde-2023-towards,zhu-etal-2024-adversarial,gkoumas-etal-2023-reformulating,li-etal-2022-gpt}, with prior work focusing mainly on prompt construction \cite{10095993,farzana-parde-2024-domain}, model choice \cite{di-palo-parde-2019-enriching}, and multi-step system design \cite{9413634,wang22l_interspeech}. Some work has leveraged speech-based approaches that extract acoustic and prosodic features to distinguish AD patients from healthy controls \cite{Ding2024,El-speech}, while others have used speech–language hybrid methods combining linguistic and acoustic cues to better capture cognitive decline \cite{shi-speech-langauge-ad}. Other work has explored language-only LLM frameworks that analyze discourse structure and lexical patterns using pretrained models \cite{zhang-2025-llm-ad,hou-2025-llm-agent,li-2025-multi-agent-llm} and vision–language models for multimodal detection \cite{Dhinagar-2025-moe}.  

In general, these studies emphasize system design and performance, with the aim of gradually improving AD detection using standardized metrics and benchmark datasets.  They have paid less attention to how AD-related information is encoded within model representations.  Our work seeks to fill this gap.

\subsection{Supervised Fine-tuning and Linear Probing}
SFT is widely used for adapting pretrained LLMs to downstream tasks by training them on task-specific labeled examples. Early work demonstrated that fine-tuning pretrained Transformers can yield substantial performance gains across NLP benchmarks compared with training models from scratch \cite{devlin2019bert,radford2019language}. Subsequent research refined SFT techniques to improve generalization, stability, and data efficiency, including by tuning instruction prompts to align models with human-written prompts and task formats \cite{chung2022scaling,wei2022instruction}. SFT has also been used as a core component in multi-stage alignment pipelines, where models are first fine-tuned on curated supervised datasets before further improvement via reinforcement learning from human feedback (RLHF) or preference optimization \cite{ouyang2022training,bai2022training}. Recent work further explores domain-specific SFT for specialized applications, such as biomedical and clinical language tasks, demonstrating that carefully selected, high-quality supervision can enable strong downstream adaptation even in low-resource settings \cite{tran2024bioinstruct, singhal2023gpt4}.

To better understand how supervised fine-tuning shapes internal representations, we next turn to probing as a lightweight and interpretable analysis tool. Linguistic and stylistic concepts often manifest as linear features within the high-dimensional spaces of language model representations. Linguistic and stylistic concepts often manifest as linear features within the high-dimensional spaces of language model representations \cite{mikolov-etal-2013-linguistic, nanda-etal-2023-emergent, pmlr-v235-park24c, gurnee-2024-space-time,kim-2025-political}. In our work, we use linear probes to estimate the direction of AD in the model’s representations and to measure the extent to which AD-related information is captured at the token level.

%% file: 3.methodology.tex
\section{Methodology}

\begin{figure*}[t]
    \includegraphics[width=0.97\linewidth]{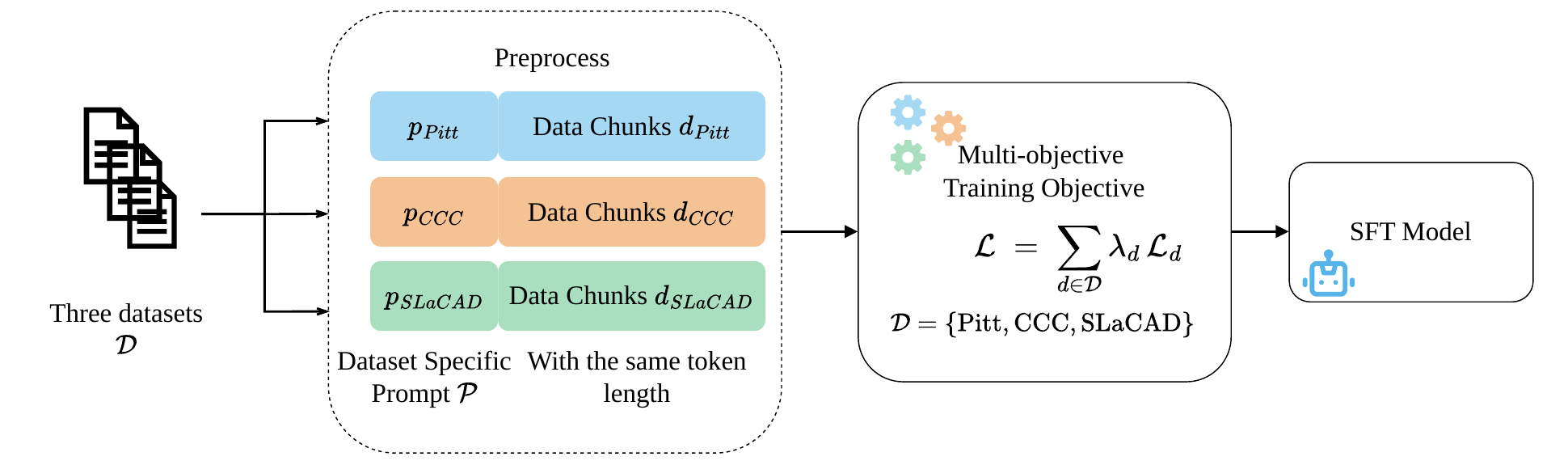} \hfill
  \caption {The pipeline of multi-objective fine-tuning for language models.}
\label{fig:trasncript-rep}
\end{figure*}

In this work, we first explore the performance of SFT with LLMs for AD detection. Additionally, we investigate how the concept of AD is encoded in LLMs' internal representations. Building on evidence that concepts are linearly encoded in representation and can be detected, we train linear probes on LLM hidden states to predict AD-specific labels, yielding probes that isolate features in the representations relevant to AD. Using these probes, we examine token-level activations in AD transcripts and identify the tokens that are most informative for downstream AD classification when processed by the models.

\subsection{Multi-Loss Learning across Heterogeneous Corpora}
\label{sec:multiloss}

As shown in Table~\ref{tab:corpora-description}, our three corpora, DementiaBank's Pitt Corpus \cite{pitt_corpus}, Carolinas Conversation Collection (CCC) \cite{ccc_corpus}, and SLaCAD \cite{farzana-etal-2024-slacad}, share the same diagnostic target
(AD vs.\ Control) but differ markedly in transcription convention, elicitation protocol, and class balance. We phrase the entire fine-tuning objective as a sum of per-corpus losses, each conditioned on a corpus-specific prompt template and a corpus-specific class weighting scheme. The same formulation is instantiated across three model families: encoder-only (BERT), encoder--decoder (T5), and decoder-only LLMs (Llama-1B).

Through the remainder of this discussion, we use the following formalisms.  We let $\mathcal{D} = \{\text{Pitt}, \text{CCC}, \text{SLaCAD}\}$.  Correspondingly, we let
$\mathcal{T}_d = \{(x_i, y_i)\}_{i=1}^{N_d}$ denote the training split of
corpus $d \in \mathcal{D}$, with binary label $y_i \in \{\text{Control},
\text{AD}\}$ shared across $\mathcal{D}$ and $d(i)$ the corpus of example $i$.

\paragraph{Corpus Conditional Prompt Templates.}
DementiaBank transcripts follow the CHAT/CLAN convention and embed annotation markers
(e.g., retracing \texttt{[/]}, revision \texttt{[//]}, replacement
\texttt{[:~\ldots]}, filled pauses \texttt{\&-uh}, \texttt{\&-um}) that are
themselves linguistically informative cues for AD detection. A more detailed example is shown in Appendix~\ref{sec:pitt-sample}. CCC is composed
of cleaned conversational interviews with no such markup, and SLaCAD contains
structured language-test recordings (autobiographical interview, normal
read-aloud, and autocorrect-cloze paragraph) that are also marker-free. 

To preserve this corpus-specific information, we use a
distinct user prompt $P_d(\cdot)$ for each corpus. The Pitt prompt embeds a
glossary of CHAT/CLAN markers together with marker-aware diagnostic criteria, whereas the CCC and SLaCAD prompts list only generic linguistic criteria (word-finding
difficulty, repetition, reduced coherence, and simpler or broken grammar). All three templates terminate in the same supervision slot.  This means that the gold answer
string $y \in \{\texttt{Control}, \texttt{AD}\}$ is identical across corpora.
The corpus-conditional input is
\begin{equation}
\tilde{x}_i \;=\; P_{d(i)}(x_i),
\label{eq:corpus-prompt}
\end{equation}


\paragraph{Per-Corpus Class-Weighted Aggregation.}
The three corpora have very different class priors: Pitt is AD-majority,
CCC is mildly Control-leaning, and SLaCAD is strongly Control-majority with only a handful of AD examples. A uniform mean over the merged pool would therefore steer gradient updates toward whichever class is over-represented
in the largest corpus. We associate each corpus with its own class-weight
vector $w_d = (w_{d,\text{Ctl}},\, w_{d,\text{AD}}) \in \mathbb{R}^{2}$,
fixed from the inverse class frequency of $\mathcal{T}_d$ in isolation, and
aggregate the per-example loss into a per-corpus loss by
\begin{equation}
\mathcal{L}_d
\;=\; \frac{1}{Z_d}\!\!\sum_{i\,\in\,\mathcal{T}_d}\!\!
   w_{d,y_i}\, \ell_\theta\!\big(P_d(x_i), y_i\big),
\label{eq:per-corpus-loss}
\end{equation}
 where $Z_d = \sum_{i \in \mathcal{T}_d} w_{d, y_i}$ normalizes the
weighted sum to a weighted mean, making $\mathcal{L}_d$ comparable
across corpora of different sizes and classes.

\paragraph{Joint Multi-Objective Loss.}
The full fine-tuning objective is the sum of the three per-corpus losses,
\begin{equation}
\mathcal{L} \;=\; \sum_{d \in \mathcal{D}} \lambda_d\, \mathcal{L}_d,
\qquad \lambda_d \geq 0,
\label{eq:joint-multiloss}
\end{equation}
with corpus mixing coefficients $\lambda_d$. We realize
Eq.~\eqref{eq:joint-multiloss} at the data preprocessing level by upsampling the smaller corpora to match the size of a reference corpus, which is Pitt in our situation: we repeat CCC and SLaCAD examples with replacement until each contributes as many training rows as Pitt(1,041 in our splits), so each $\mathcal{L}_d$ receives equal exposure per epoch.

\paragraph{Standard SFT (Cross-Entropy Loss).}
Standard SFT employs cross-entropy (CE) loss to maximize the likelihood of the ground-truth labels:
\begin{equation}
\mathcal{L}_{\text{CE}} = - \sum_{i=1}^{C} y_i \log(p_i),
\end{equation}
where \(y_i\) denotes the one-hot encoded ground-truth label and \(p_i\) is the predicted probability for class \(i\).

\paragraph{Focal Loss.}
To explicitly down-weight easy, confidently classified examples and
concentrate the gradient on hard or minority-class ones, we replace
$\mathcal{L}_{\text{CE}}$ with the focal cross-entropy of
\cite{}:
\begin{equation}
\mathcal{L}_{\text{focal}} \;=\; - \sum_{i=1}^{C} (1 - p_i)^{\gamma}\,
y_i \log(p_i),
\label{eq:focal-loss}
\end{equation}
where the modulating factor $(1 - p_i)^{\gamma}$ vanishes as the model
assigns high probability to the correct class, and $\gamma \geq 0$ controls
the rate of this attenuation; we fix $\gamma = 2$ throughout. 

\paragraph{Label Smoothing.}
Label smoothing \cite{} softens the one-hot target by
mixing it with a uniform distribution over the $C$ classes,
\begin{equation}
\begin{aligned}
\tilde{y}_i &\;=\; (1 - \alpha)\, y_i + \frac{\alpha}{C}, \\
\mathcal{L}_{\text{LS}} &\;=\; - \sum_{i=1}^{C} \tilde{y}_i \log(p_i),
\end{aligned}
\label{eq:label-smoothing}
\end{equation}
where $\alpha \in [0, 1)$ controls the smoothing strength; we set
$\alpha = 0.1$. Minimising $\mathcal{L}_{\text{LS}}$ instead of $\mathcal{L}_{\text{CE}}$
penalises overconfident next-token predictions on the ground truth label
(\textit{AD} or \textit{Control}). This is useful in AD-detection SFT
because the training corpora differ in task, annotation, and class
balance; Smoothing discourages the model from overfitting to
corpus-specific patterns and produces more reliable label predictions
at generation time.

\subsection{Probing}
After studying SFT performance for AD detection, we ask whether AD-relevant linguistic information is linearly recoverable from the model's internal states. Following \citet{gurnee-2024-space-time}, we fit a per-layer ridge probe $\hat{y} = \mathbf{W}^\top h$ that maps a hidden representation $h \in \mathbb{R}^d$ to the AD label $y \in \{0, 1\}$ (0 = Control, 1 = AD), with objective
\begin{equation}
\ell(\mathbf{W}) = \frac{1}{N}\sum_{i=1}^{N}\big(y_i - \mathbf{W}^\top h_i\big)^2 + \lambda_{\text{ridge}}\,\|\mathbf{W}\|_2^2,
\end{equation}
where $N$ is the number of training samples and $\lambda_{\text{ridge}}$ controls the L2 penalty. The fitted $\mathbf{W}$ defines a linear direction for AD in representation space, and $\hat{y}_i$ measures how strongly that concept is expressed in $h_i$.

%% file: 4.main_experiments.tex
\section{Experiments}
\subsection{Experimental Settings}

\paragraph{Hyperparameters.}
Across all experiments, we train for 10 epochs. The Llama-3.2-1B-Instruct model is fine-tuned with a learning rate of 2e-5, preserving long transcripts without truncation. The T5-large baseline is fine-tuned at a learning rate of 3e-4, while the BERT-base baseline uses a learning rate of 2e-5. To counter the class imbalance, which differs across corpora, we apply a per-task weighted cross-entropy loss with dataset-specific (Control, AD) weights that up-weight each corpus's minority class in inverse proportion to its frequency, using a stronger exponent on SLaCAD to absorb its extreme, inverted imbalance: (2.099, 0.477) for Pitt, (1.277, 0.783) for CCC, and (0.096, 10.429) for SLaCAD.  Label smoothing and focal loss additionally introduce a label-smoothing factor of 0.1, focal\_alpha=0.25 with focal\_gamma=2.0, and a contrastive mixing weight of 0.1 with a margin of 1.0, respectively.

\begin{table*}[t]
\centering
\setlength{\tabcolsep}{4pt}
\renewcommand{\arraystretch}{1.15}
\begin{tabular}{llcccccccc}
\toprule
& & \multicolumn{2}{c}{\textbf{Pitt}} & \multicolumn{2}{c}{\textbf{CCC}} & \multicolumn{2}{c}{\textbf{SLaCAD}} & \multicolumn{2}{c}{\textbf{Mean}} \\
\cmidrule(lr){3-4}\cmidrule(lr){5-6}\cmidrule(lr){7-8}\cmidrule(lr){9-10}
\textbf{Model} & \textbf{Loss Type} & \textbf{Acc} & \textbf{F1} & \textbf{Acc} & \textbf{F1} & \textbf{Acc} & \textbf{F1} & \textbf{Acc }& \textbf{F1} \\
\midrule
\multirow{4}{*}{Llama-1B}
& Vanilla         & 0.78{\tiny$\pm$0.00} & 0.44{\tiny$\pm$0.00} & 0.61{\tiny$\pm$0.00} & 0.50{\tiny$\pm$0.00} & 0.26{\tiny$\pm$0.00} & 0.24{\tiny$\pm$0.00} & 0.55 & 0.39 \\
 & Standard        & 0.84{\tiny$\pm$0.01} & 0.73{\tiny$\pm$0.02} & 0.74{\tiny$\pm$0.02} & 0.70{\tiny$\pm$0.02} & 0.87{\tiny$\pm$0.05} & 0.46{\tiny$\pm$0.01} & 0.82 & 0.63 \\
 & Weighted        & \textbf{0.88}{\tiny$\pm$0.02} & \textbf{0.83}{\tiny$\pm$0.02} & \textbf{0.83}{\tiny$\pm$0.02} & \textbf{0.82}{\tiny$\pm$0.02} & \textbf{0.89}{\tiny$\pm$0.02} & \textbf{0.63}{\tiny$\pm$0.02} & \textbf{0.87} & \textbf{0.76} \\
 & Focal           & 0.77{\tiny$\pm$0.02} & 0.68{\tiny$\pm$0.02} & 0.63{\tiny$\pm$0.02} & 0.62{\tiny$\pm$0.02} & 0.78{\tiny$\pm$0.09} & 0.44{\tiny$\pm$0.03} & 0.73 & 0.58 \\
 & Label smoothing & 0.79{\tiny$\pm$0.01} & 0.75{\tiny$\pm$0.02} & 0.69{\tiny$\pm$0.04} & 0.69{\tiny$\pm$0.04} & 0.69{\tiny$\pm$0.02} & 0.48{\tiny$\pm$0.01} & 0.73 & 0.64 \\
\midrule
T5-large  & Standard & 0.92 & 0.89 & 0.96 & 0.95 & 0.97 & 0.83 & 0.95 & 0.89 \\
BERT-base & Standard & 0.91 & 0.87 & 0.93 & 0.92 & 0.95 & 0.49 & 0.93 & 0.76 \\
\bottomrule
\end{tabular}%
\caption{Accuracy and macro-F1 on Pitt, CCC, and SLaCAD. T5/BERT are single deterministic runs. "Standard" denotes uniform per-task, per-class loss weights, all set to 1.
}
\label{tab:loss_type_comparison}

\end{table*}

\paragraph{Other Modeling Details.}
We experiment with each loss function condition on both backbone models across three corpora. The DementiaBank corpus consists of 1{,}044 AD and 247 Control samples; the CCC corpus consists of 57 AD  and 48 Control samples; and the SLaCAD corpus consists of 9 AD and 82 Control samples, exhibiting the opposite class imbalance to DementiaBank. Each corpus is split 80/20 into training and evaluation sets in a stratified manner.
We conducted all experiments using four A100 PCIe GPUs, each equipped with 40 GB of memory. Fine-tuning a single model on 4 GPUs takes approximately 6 minutes.

\subsection{Supervised Fine-tuning Results}

Table~\ref{tab:loss_type_comparison} compares different loss functions for Llama-1B across the Pitt, CCC, and SLaCAD datasets, using Accuracy (Acc) and F1-score as evaluation metrics. Among all investigated loss functions, the \textbf{Weighted loss} consistently achieved the best performance across all datasets. Specifically, it obtained the highest mean Accuracy ($0.87$) and mean F1-score ($0.76$), outperforming the Standard, Focal, and Label Smoothing losses.

For the Pitt dataset, the Weighted loss improved the F1-score from $0.73$ (Standard loss) to $0.83$, indicating a substantial gain in classification balance and robustness. Similarly, on the CCC dataset, the Weighted loss achieved the best Accuracy ($0.83$) and F1-score ($0.82$), demonstrating its effectiveness in handling class imbalance. The largest relative improvement was observed on the SLaCAD dataset, where the F1-score increased from $0.46$ under Standard loss to $0.63$ using Weighted loss.

In contrast, the Focal loss consistently underperformed compared to the other approaches, yielding the lowest mean Accuracy ($0.73$) and mean F1-score ($0.58$). Label smoothing provided moderate improvements over Focal loss but did not surpass the Standard or Weighted configurations.

Compared with larger baseline models, T5-large achieved the best overall performance, with a mean Accuracy of $0.95$ and a mean F1-score of $0.89$. BERT-base also outperformed Llama-1B in most settings, particularly on the Pitt and CCC datasets. However, the Weighted-loss Llama-1B significantly narrowed the performance gap, suggesting that appropriate loss design can substantially improve the performance of smaller language models.

\paragraph{Summary of SFT Results}
Overall, the experimental results demonstrate that applying a Weighted loss function is the most effective strategy for improving Llama-1B performance across all evaluated datasets. The consistent gains in both Accuracy and F1-score suggest that class imbalance is a critical factor in these tasks, and weighting the loss successfully mitigates this issue. Although larger models such as T5-large still achieve superior overall results, the optimized Llama-1B configuration offers a competitive and computationally efficient alternative.

\paragraph{Impact of Class Imbalance on SLaCAD.}
The SLaCAD dataset is highly imbalanced, containing only $7$ Alzheimer’s disease (AD) samples while the remaining subjects belong to the control class. This imbalance explains the discrepancy between Accuracy and F1-score, as models can achieve high Accuracy by favoring the majority class while failing to correctly identify AD samples. For example, although Standard loss achieved relatively high Accuracy on SLaCAD, the corresponding F1-scores remained low. In contrast, the Weighted loss substantially improved the F1-score, indicating better recognition of the minority AD class and demonstrating the importance of class-balanced optimization for highly skewed datasets.

%% file: 5.additional.tex
\section{Ablation Studies}
\label{sec:ablations}

\begin{table*}[t]
\centering

\setlength{\tabcolsep}{4pt}
\begin{tabular}{l cc cc cc}
\toprule
& \multicolumn{2}{c}{\textbf{Pitt}}
& \multicolumn{2}{c}{\textbf{CCC}}
& \multicolumn{2}{c}{\textbf{SLaCAD}} \\
\cmidrule(lr){2-3}\cmidrule(lr){4-5}\cmidrule(lr){6-7}
\textbf{Active Loss Terms} & \textbf{Acc} & \textbf{F1} & \textbf{Acc} & \textbf{F1} & \textbf{Acc} & \textbf{F1} \\
\midrule
Pitt + CCC + SLaCAD
  & $\mathbf{0.88 {\scriptstyle\,\pm\,0.02}}$ & $\mathbf{0.83 {\scriptstyle\,\pm\,0.02}}$
  & $0.83 {\scriptstyle\,\pm\,0.02}$          & $0.82 {\scriptstyle\,\pm\,0.02}$
  & $0.89 {\scriptstyle\,\pm\,0.02}$          & $\mathbf{0.63 {\scriptstyle\,\pm\,0.02}}$ \\
Pitt + CCC
  & $0.85 {\scriptstyle\,\pm\,0.01}$          & $0.79 {\scriptstyle\,\pm\,0.02}$
  & $\mathbf{0.85 {\scriptstyle\,\pm\,0.03}}$ & $\mathbf{0.84 {\scriptstyle\,\pm\,0.03}}$
  & $0.76 {\scriptstyle\,\pm\,0.03}$          & $0.49 {\scriptstyle\,\pm\,0.05}$ \\
Pitt + SLaCAD
  & $0.86 {\scriptstyle\,\pm\,0.01}$          & $0.81 {\scriptstyle\,\pm\,0.02}$
  & $0.66 {\scriptstyle\,\pm\,0.02}$          & $0.57 {\scriptstyle\,\pm\,0.03}$
  & $\mathbf{0.90 {\scriptstyle\,\pm\,0.02}}$ & $0.47 {\scriptstyle\,\pm\,0.00}$ \\
CCC + SLaCAD
  & $0.68 {\scriptstyle\,\pm\,0.01}$          & $0.48 {\scriptstyle\,\pm\,0.02}$
  & $0.64 {\scriptstyle\,\pm\,0.01}$          & $0.42 {\scriptstyle\,\pm\,0.03}$
  & $0.17 {\scriptstyle\,\pm\,0.07}$          & $0.16 {\scriptstyle\,\pm\,0.06}$ \\
\midrule
Pitt only
  & $0.86 {\scriptstyle\,\pm\,0.00}$          & $0.80 {\scriptstyle\,\pm\,0.01}$
  & $0.57 {\scriptstyle\,\pm\,0.01}$          & $0.53 {\scriptstyle\,\pm\,0.02}$
  & $0.57 {\scriptstyle\,\pm\,0.11}$          & $0.42 {\scriptstyle\,\pm\,0.04}$ \\
CCC only
  & $0.71 {\scriptstyle\,\pm\,0.02}$          & $0.48 {\scriptstyle\,\pm\,0.04}$
  & $0.56 {\scriptstyle\,\pm\,0.05}$          & $0.49 {\scriptstyle\,\pm\,0.07}$
  & $0.44 {\scriptstyle\,\pm\,0.05}$          & $0.33 {\scriptstyle\,\pm\,0.02}$ \\
SLaCAD only
  & $0.54 {\scriptstyle\,\pm\,0.01}$          & $0.49 {\scriptstyle\,\pm\,0.00}$
  & $0.46 {\scriptstyle\,\pm\,0.07}$          & $0.45 {\scriptstyle\,\pm\,0.08}$
  & $0.75 {\scriptstyle\,\pm\,0.05}$          & $0.54 {\scriptstyle\,\pm\,0.03}$ \\
\bottomrule
\end{tabular}
\caption{
Loss-term ablation for Llama-1B. Pitt + CCC + SLaCAD is the full version with all three loss terms.
}
\label{tab:loss_ablation}

\end{table*}
We ablate three design choices that together specify the multi-loss objective
in Eq.~\eqref{eq:joint-multiloss}: (i) which per-corpus loss terms
$\mathcal{L}_d$ are active, (ii) how each corpus is preprocessed before being
fed to the model, and (iii) whether the corpus-conditional prompt $P_d$ for
Pitt should preserve its CHAT/CLAN markers. Unless otherwise noted, every
ablation uses the same backbone, the same per-corpus class weights $w_d$, the
same upsample-to-reference mixing policy ($\lambda_d = 1$), and the same
held-out test splits; only the variable under study changes. We report
per-corpus macro-$F_1$ on Pitt, CCC, and SLaCAD and the mean macro-$F_1$ across
corpora. 

\begin{table*}[t]
\centering

\setlength{\tabcolsep}{4pt}
\begin{tabular}{l cc cc cc}
\toprule
& \multicolumn{2}{c}{\textbf{Pitt}}
& \multicolumn{2}{c}{\textbf{CCC}}
& \multicolumn{2}{c}{\textbf{SLaCAD}} \\
\cmidrule(lr){2-3}\cmidrule(lr){4-5}\cmidrule(lr){6-7}
\textbf{Chunk Size} & \textbf{Acc} & \textbf{F1} & \textbf{Acc} & \textbf{F1} & \textbf{Acc} & \textbf{F1} \\
\midrule
1024
  & $0.83 {\scriptstyle\,\pm\,0.01}$ & $0.69 {\scriptstyle\,\pm\,0.03}$ & $0.82 {\scriptstyle\,\pm\,0.01}$ & $0.79 {\scriptstyle\,\pm\,0.02}$ & $0.86 {\scriptstyle\,\pm\,0.03}$ & $0.46 {\scriptstyle\,\pm\,0.01}$ \\
512        & \textbf{0.88}{\tiny$\pm$0.02} & \textbf{0.83}{\tiny$\pm$0.02} & \textbf{0.83}{\tiny$\pm$0.02} & \textbf{0.82}{\tiny$\pm$0.02} & \textbf{0.89}{\tiny$\pm$0.02} & \textbf{0.63}{\tiny$\pm$0.02}\\
\bottomrule

\end{tabular}

\caption{%
Chunk-size ablation for the  Llama-1B model.
}
\label{tab:chunk_ablation}

\end{table*}

\subsection{Per-Corpus Loss-Term Ablation}
\label{sec:abl-loss-terms}

This study tests whether the three per-corpus losses are complementary or
whether a single dominant corpus accounts for the model's behavior. We enable
every non-empty subset $\mathcal{S} \subseteq \{\text{Pitt}, \text{CCC},
\text{SLaCAD}\}$ in turn and train with
$\mathcal{L}^{(\mathcal{S})} = \sum_{d \in \mathcal{S}} \mathcal{L}_d$,
holding all other components fixed. Concretely, in each variant, the corpora
in $\mathcal{S}$ contribute their full per-corpus weighted cross-entropy from
Eq.~\eqref{eq:per-corpus-loss}, while the remaining corpora are removed from
the training mix entirely. Table~\ref{tab:loss_ablation} reports the seven
resulting models.

\paragraph{Findings.}\textbf{ All corpora contribute complementary signal.}
Activating all three losses yields the best results across all per-corpus metrics. The single-corpus models confirm that Pitt is the only
corpus whose training signal alone produces a reasonable Pitt $F_1$ (0.804),
but it still leaves both CCC and SLaCAD near majority-class performance.
Crucially, removing Pitt from the loss (CCC+SLaCAD)
collapses every metric, including SLaCAD's own $F_1$; this shows
that the abundant, AD-rich Pitt signal acts as an anchor through which the
smaller corpora's weighted terms become learnable. The result supports
keeping all three $\mathcal{L}_d$ in the joint objective rather than
deferring CCC or SLaCAD to a downstream fine-tune.

\subsection{Chunking}
\label{sec:abl-chunk}

Table~\ref{tab:chunk_ablation} shows that reducing the chunk size from 1024 to 512 consistently improves performance across all datasets for the Llama-1B model. On the Pitt dataset, accuracy increases from 0.83 to 0.88, while the F1 score improves substantially from 0.69 to 0.83. Similar gains are observed on the CCC dataset, where accuracy rises slightly from 0.82 to 0.83, and F1 improves from 0.79 to 0.82. The largest improvement is seen on the SLaCAD dataset: accuracy increases from 0.86 to 0.89, and F1 scores from 0.46 to 0.63. Overall, the results indicate that a smaller chunk size of 512 yields more robust, balanced classification performance, particularly in terms of F1 score.

\paragraph{Findings.} \textbf{Chunk size should match the shortest corpus.}
The intuition behind using a chunk size of 512 is that it creates text segments with more uniform and comparable token lengths across data points, refers to Figure~\ref{fig:token-length}. We argue that a similar data segment length can help the model understand this domain better.

\subsection{Corpus-Pair Composition and the Role of CHAT/CLAN Markers}
\label{sec:abl-pair}

\begin{table}[t]
\centering
\setlength{\tabcolsep}{3pt}
\renewcommand{\arraystretch}{1.05}
\begin{tabular}{L{2.5cm} cc cc}
\toprule
 & \multicolumn{2}{c}{\textbf{Pitt}} & \multicolumn{2}{c}{\textbf{CCC}} \\
\cmidrule(lr){2-3} \cmidrule(lr){4-5}
\textbf{Training Data} & \textbf{Acc} & \textbf{F1} & \textbf{Acc} & \textbf{F1} \\
\midrule
Pitt only            & .85\textsubscript{\,.01} & .74\textsubscript{\,.03} & .60\textsubscript{\,.02} & .45\textsubscript{\,.01} \\
CCC only             & .73\textsubscript{\,.02} & .50\textsubscript{\,.02} & .61\textsubscript{\,.03} & .44\textsubscript{\,.02} \\
Pitt+CCC, w/ markers & \textbf{.85}\textsubscript{\,.01} & \textbf{.79}\textsubscript{\,.02} & \textbf{.85}\textsubscript{\,.03} & \textbf{.84}\textsubscript{\,.03} \\
Pitt+CCC, w/o markers & \textbf{.85}\textsubscript{\,.01} & .78\textsubscript{\,.01} & .84\textsubscript{\,.01} & .82\textsubscript{\,.01} \\
\bottomrule
\end{tabular}
\caption{Marker ablation.}
\label{tab:marker_ablation}
\end{table}

In this experiment, we investigate whether the Pitt-specific CHAT/CLAN markers
in $P_{\text{Pitt}}$ help or hurt cross-corpus transfer. We test both
questions at once on the Pitt+CCC pair, which is the only pair in which both corpora are sufficiently large for a stratified test split. We compare
four configurations: each corpus trained alone with its own prompt; a
balanced Pitt+CCC mix using the canonical $P_{\text{Pitt}}$ that exposes
CHAT/CLAN markers, and the same balanced mix with Pitt markers stripped
and Pitt fed through the no-marker prompt variant
(Section~\ref{sec:multiloss}).

\paragraph{Findings.} \textbf{Pitt's CHAT/CLAN markers are a transferable AD-domain feature.}
First, the balanced Pitt+CCC mix beats both
single-corpus models by a wide margin: mean $F_1$ improves by $+0.22$
over Pitt-only and $+0.35$ over CCC-only, with each joint model gaining
most on the corpus it had \emph{not} previously seen (CCC $F_1$ rises
from $0.44$ to $0.84$ once Pitt is added). This matches the loss-term
ablation.  Per-corpus losses contribute a non-redundant signal, and the only way to fit both well is to learn a representation invariant to
surface differences between the two corpora. 

Second, keeping the CHAT/CLAN markers yields a small but consistent gain over the no-marker variant ($+0.01$ Pitt $F_1$, $+0.02$ CCC $F_1$, accuracies essentially
unchanged). Markers such as \texttt{\&-um}, \texttt{[/]}, and
\texttt{+...} are unique to Pitt's transcription convention and
explicitly tag the disfluencies, retraces, and word-finding pauses that characterize AD speech but are not annotated in any other corpus we train on; the markers therefore expose a small AD-domain feature the model picks up from Pitt and reuses on unmarked CCC, where the same phenomena are signaled only lexically.

\subsection{Summary of Ablations}
\label{sec:abl-summary}

The three ablations converge on a single recommended configuration.
\textbf{(1) All corpora contribute complementary signal.}
Activating every per-corpus loss $\mathcal{L}_d$ in the joint objective
yields the best score on every per-corpus metric; dropping any single
$\mathcal{L}_d$, and most dramatically $\mathcal{L}_{\text{Pitt}}$,
degrades performance on the remaining corpora as well as on the
dropped one.
\textbf{(2) Chunk size should match the shortest corpus.}
Setting the chunking budget to the length scale of the smallest representative segment gives all training examples a
comparable input granularity and improves $F_1$ uniformly across all
three corpora.
\textbf{(3) Pitt's CHAT/CLAN markers are a transferable AD-domain feature.}
The diverse marker vocabulary in $P_{\text{Pitt}}$ explicitly tags
disfluencies, retraces, and word-finding pauses that other corpora
leave unannotated; retaining it improves generalization on
\emph{both} Pitt and unmarked CCC, indicating that the model extracts
a small but genuine AD-domain signal from these markers and reuses it
on inputs that lack them.

\subsection{Probing Representation}

We trained probes by using hidden states from the SOTA Llama-1B to test how well dementia-vs-control information is recoverable from each layer's representation. Each probe is a ridge-regularized linear head on top of the frozen last-token hidden state at the target layer; the underlying language model is kept frozen. Because SLaCAD's extreme class imbalance ($\approx$11:1) makes per-layer probe metrics unstable, we restrict probing to the Pitt and CCC training splits. The results are shown in \ref{fig:probing-training}. We chose layer-15 as the probe according to the 4 metrics

\begin{figure}[t]
    \includegraphics[width=0.95\linewidth]{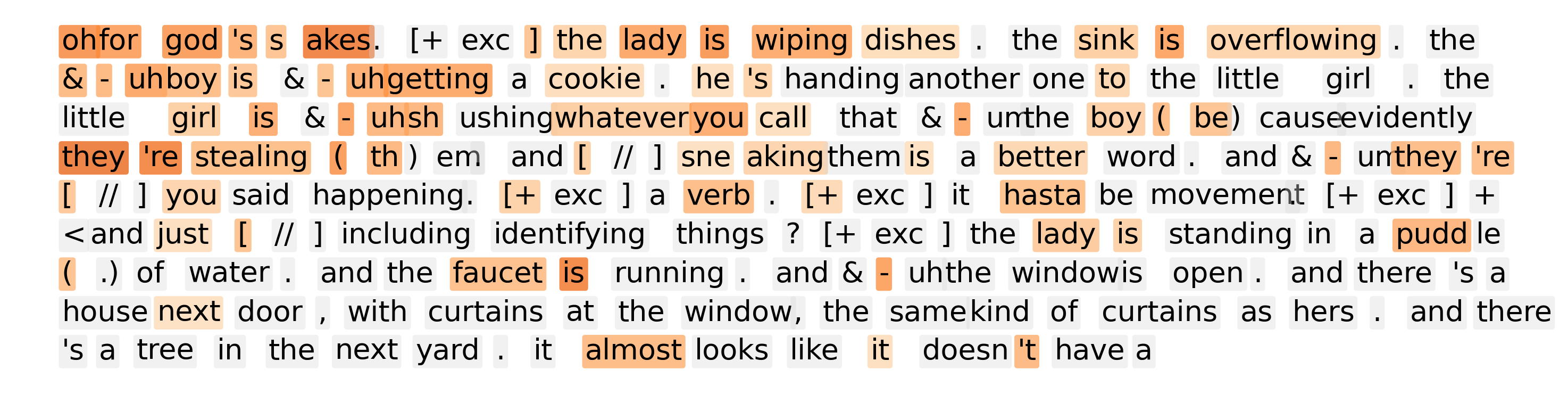}
    \includegraphics[width=0.95\linewidth]{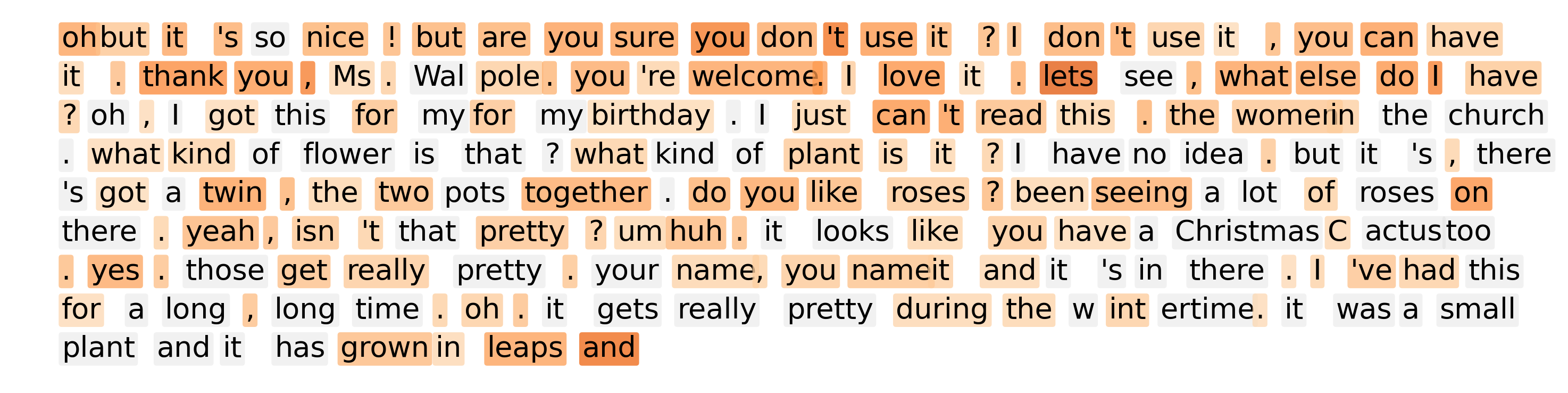}
  \caption {This figure illustrates the differences in token probing values before and after fine-tuning of the transcripts. The upper panel corresponds to the Pitt dataset, while the lower panel corresponds to the CCC dataset.}
\label{fig:trasncript-rep}
\end{figure}

Figure \ref{fig:trasncript-rep} illustrates our token-level representation analysis on two example transcripts, showing how fine-tuning reshapes the AD-related signal carried by each token. For every token, we project its layer-15 hidden state onto the linear probe direction learned from the base and fine-tuned models, and visualize the signed difference; color intensity encodes the magnitude of change, with cells below a threshold of 0.02 left unshaded. Two complementary patterns emerge. On the picture-description sample, fine-tuning amplifies the signal on CHAT-format disfluency and revision markers (\texttt{\&-uh}, \texttt{\&-un}, \texttt{[//]}, \texttt{[+ exc]}, partial-word fragments such as \texttt{sne}aking and \texttt{pudd}le), together with scene-grounded content words (\textit{lady}, \textit{sink}, \textit{wiping}, \textit{overflowing}, \textit{stealing}, \textit{faucet}) and affective exclamations (\textit{oh for god's sakes}). On the conversational sample, the largest shifts concentrate on interactional fillers and backchannels (\textit{oh}, \textit{yeah}, \textit{um huh}, \textit{yes}), repetitive or parallel constructions indicative of perseveration and circumlocution 
, and high-frequency pronouns and formulaic question scaffolds.

Together, these examples show that the probe isolates a latent representational direction along which AD-related linguistic variation is expressed in the base language model, and that token-level projections along this direction pinpoint the lexical, disfluency, and discourse-level elements the model learns to emphasize or de-emphasize during fine-tuning, yielding interpretable, fine-grained evidence of \textit{which} linguistic phenomena drive the classifier. 

%% file: 6.conclusion.tex
\section{Conclusion}

This work demonstrates that a small, open-weight LLM, when trained with a multi-loss supervised fine-tuning framework, can achieve strong, consistent Alzheimer’s disease detection performance across heterogeneous clinical speech corpora. Our results show that jointly leveraging complementary datasets, carefully selecting chunk granularity, and preserving clinically meaningful transcript annotations together improve robustness and generalization. Beyond establishing competitive SOTA performance, this study provides an important foundation for future exploration of generative LLMs as scalable and transferable tools for multi-corpus clinical AD assessment workflows.

%% file: 7.limitation.tex
\section{Limitations}

A limitation of this study is the strong class imbalance in SLaCAD, which may affect the stability and generalizability of the reported performance despite the robustness of the proposed multi-loss framework. In addition, although our approach demonstrates strong results across Pitt, CCC, and SLaCAD, the pipeline should be further evaluated on additional external and more diverse datasets to better assess its robustness, transferability, and clinical applicability across different populations, recording conditions, and transcript styles.

Another limitation is that this work primarily focuses on classification performance rather than interpretability or reasoning. While LLMs possess advanced contextual and reasoning capabilities, understanding how the model connects specific linguistic patterns in transcripts to its final AD detection decisions remains largely unexplored. Establishing more transparent reasoning pathways and clinically interpretable evidence for model predictions is an important direction for future research.

%% file: 8.appendix.tex
\section{Special Markers}
\label{sec:pitt-sample}
\begin{table*}[h!]
\centering

\begin{tabular}{l|l|l}
\toprule
\textbf{Symbol / Mark} & \textbf{Meaning} & \textbf{Example} \\
\midrule
\texttt{*PAR:} / \texttt{*INV:} & Speaker label & \texttt{*PAR:} = participant, \texttt{*INV:} = interviewer \\
\texttt{[...]} & Unintelligible speech & I went to the [...] yesterday. \\
\texttt{xxx} & Unintelligible word & He xxx the cookies. \\
\texttt{?} & Uncertain transcription & I saw a dog? \\
\texttt{((...))} & Non-verbal actions & ((laughs)), ((coughs)) \\
\texttt{=} & Repetition / continuation & I = I went there. \\
\texttt{\%mor:} & Morphosyntactic tier & pro|The n|boy v|be \\
\texttt{\%com:} & Comment / notes & Optional researcher annotations \\
\bottomrule
\end{tabular}
\caption{Common CHAT special marks in Pitt transcripts.}
\label{tab:chat_marks}
\end{table*}

A short excerpt from a Pitt transcript illustrates these conventions:

\begin{verbatim}
*PAR: The boy is = is taking cookies.
%mor: pro|The n|boy v|be aux|is v|take 
n|cookie
*INV: What is he doing?
*PAR: ((points to the picture)) I don't 
know.
*PAR: He uh uh ((laughs)) is taking 
cookies?
\end{verbatim}

In this example, \texttt{*PAR:} and \texttt{*INV:} denote the participant and interviewer, \texttt{=} indicates repetition, \texttt{uh uh} represents hesitation, and \texttt{((laughs))} captures nonverbal behavior. Such detailed annotations allow researchers to analyze speech patterns, lexical retrieval, and cognitive markers, which are particularly relevant in studies of dementia.

\section{Linguistic Marker Set}

\begin{table*}[h]
    \centering

    \begin{tabular}{cccc}
        \toprule
        \textbf{Pattern} & \textbf{Regex} & \textbf{Example Markers} & \textbf{Description} \\
        \midrule
        1 & \verb|&-\w+| & \verb|&-uh|, \verb|&-um| & Filled pauses \\
        2 & \verb|&=\w+[:\w]*| & \verb|&=clears_throat|, \verb|&=sighs| & Non-verbal sounds \\
        3 & \verb|\[\+\s*[^\]]*\]| & \verb|[+ gram]|, \verb|[+ exc]| & Grammatical/exclamation \\
        4 & \verb|\[/?/?\]| & \verb|[/]|, \verb|[//]| & Retracing \\
        5 & \verb|\[:\s*[^\]]*\]| & \verb|[: word]|, \verb|[: ...]| & Replacement \\
        6 & \verb|<[^>]*>| & \verb|<...>|, \verb|<word>| & Uncertain/omitted \\
        7 & \verb|\[[^\]]*\]| & \verb|[word]|, \verb|[xxx]| & Any brackets \\
        8 & \verb|\+\<|\verb!+\>!| & \verb|+<|, \verb|+>| & Other markers \\
        9 & \verb|\(\.+\)| & \verb|(.)|, \verb|(..)|, \verb|(...)| & Pause markers \\
        10 & \verb|\bxxx\b| & \verb|xxx| & Unintelligible \\
        \bottomrule
    \end{tabular}
    \caption{Linguistic marker set.}
    \label{tab:marker-set}
\end{table*}

\section{The Token Length Distribution of Each Dataset}

\begin{figure}[h]
    \includegraphics[width=0.95\linewidth]{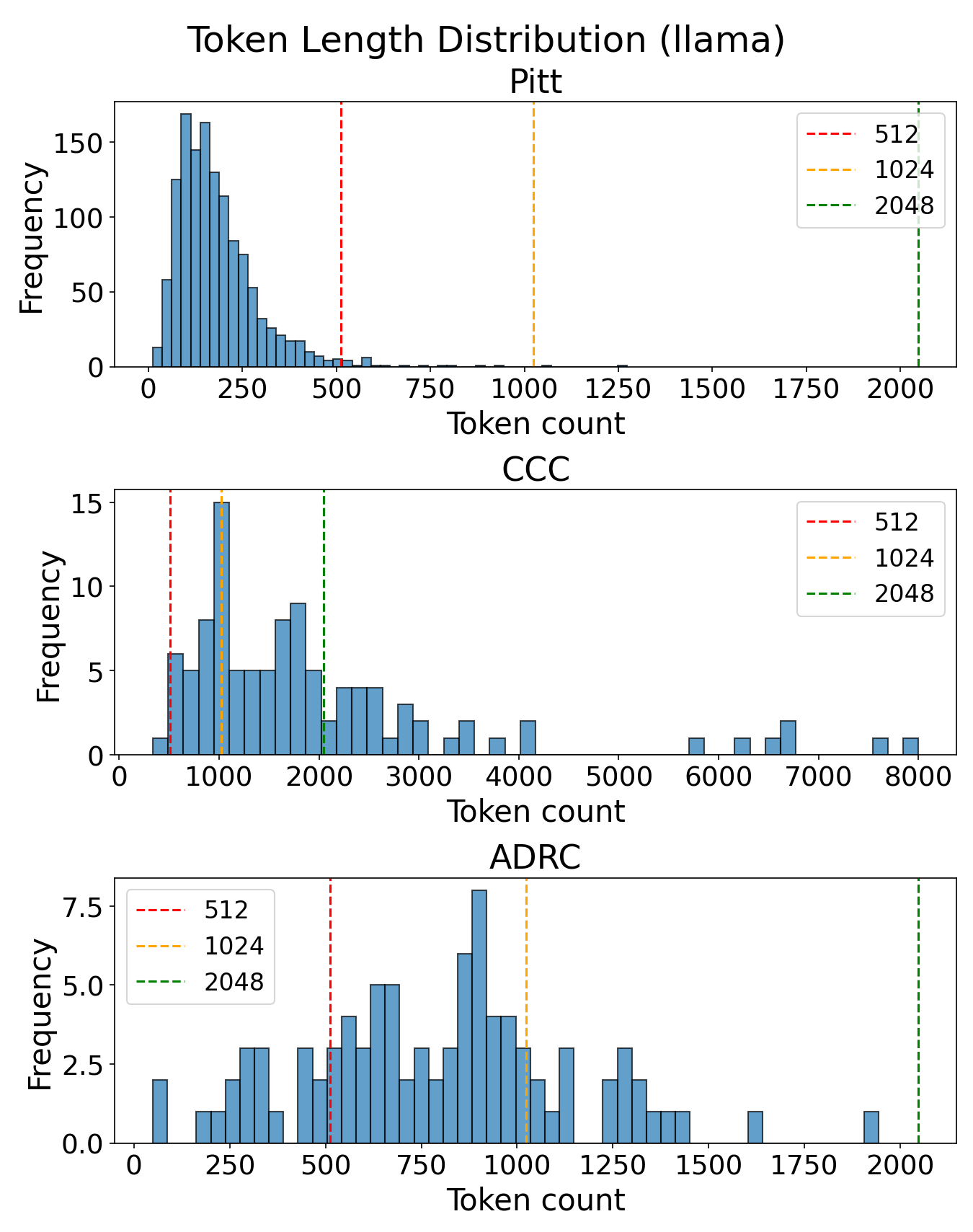}
  \caption {This figure shows the token length distribution of each dataset with the Llama3.2-1B-Instruct tokenizer.}
  \label{fig:token-length}
\end{figure}

\section{Probing Training}
\begin{figure}[h]
    \includegraphics[width=0.95\linewidth]{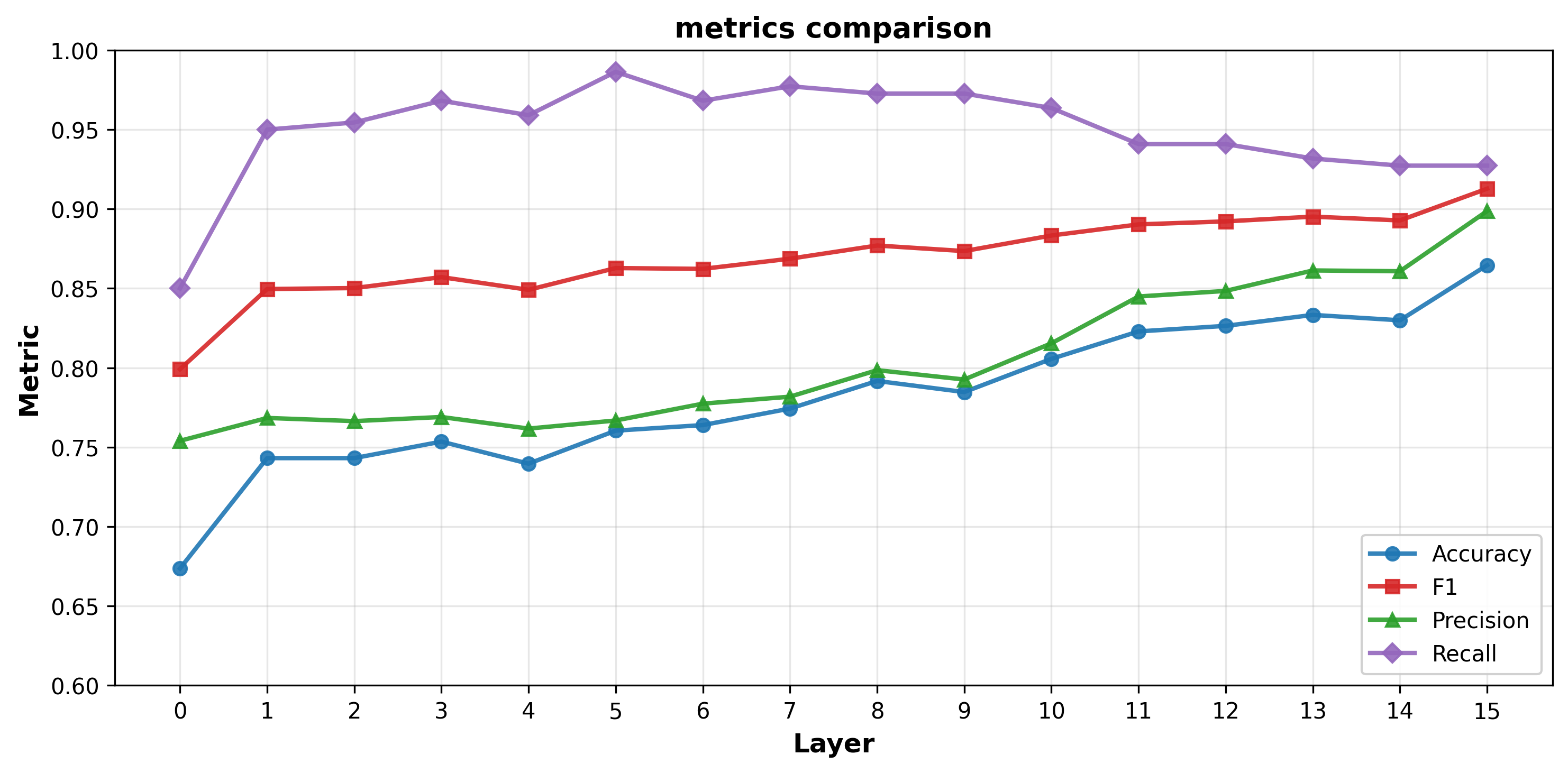}
  \caption {This figure shows probing training results of the SFT Llama3.2-1B-Instruct model.}
  \label{fig:probing-training}
\end{figure}